\documentclass[manuscript]{acmart}

\AtBeginDocument{%
  }

\copyrightyear{2022}
\acmYear{2022}
\setcopyright{acmlicensed}\acmConference[HT '22]{Proceedings of the 33rd ACM Conference on Hypertext and Social Media}{June 28-July 1, 2022}{Barcelona, Spain}
\acmBooktitle{Proceedings of the 33rd ACM Conference on Hypertext and Social Media (HT '22), June 28-July 1, 2022, Barcelona, Spain}
\acmPrice{15.00}
\acmDOI{10.1145/3511095.3531276}
\acmISBN{978-1-4503-9233-4/22/06}

\acmJournal{TOG}
\acmVolume{37}
\acmNumber{4}
\acmArticle{111}
\acmMonth{8}

\usepackage{amsmath}
\usepackage{graphicx}
\usepackage{booktabs}
\usepackage{listings}
\usepackage{xspace}
\usepackage{algorithm,algorithmic}
\usepackage{microtype}
\usepackage{hyperref}

\usepackage{bbm}
\usepackage[nolist]{acronym}
\begin{acronym}[UML]
	\acro{AI}{Artificial Intelligence}
	\acro{BFS}{Breadth-First-Search}
	\acro{BoW}{Bag-of-Words}
	\acro{CSV}{Comma-Separated Values}
	\acro{CNN}{Convolutional Neural Network}
	\acro{DL}{Description logic}

	\acro{ER}{Entity Resolution}
	\acro{EM}{Expectation Maximization}
	\acro{EBMT}{Example-Based Machine Translation}
	\acro{EBNF}{Extended Backus--Naur Form}
	\acro{EL}{Entity Linking}
	\acro{FAO}{Food and Agriculture Organization of the United Nations}
	\acro{GIS}{Geographic Information Systems}
	\acro{GHO}{Global Health Observatory}
	\acro{GRU}{Gated recurrent unit}
	\acro{HDI}{Human Development Index}
	\acro{ICT}{Information and communication technologies}
	\acro{IFRS}{International Financial Reporting Standards}
	\acro{ICD}{International Classification of Diseases}
	\acro{IT}{Information Technology}
    \acro{KB}{Knowledge Base}
    \acro{KG}{Knowledge Graph}
    \acrodefplural{KG}{Knowledge Graphs}
    \acro{KGE}{Knowledge Graph Embeddings}
    \acro{KBSE}{Knowledge Base Semantic Embedding}
    \acro{KGC}{Knowledge Graph Completion}
	\acro{LR}  {Language Resource}
	\acro{LD}  {Linked Data}
	\acro{LLOD}  {Linguistic Linked Open Data}
	\acro{LIMES}{LInk discovery framework for MEtric Spaces}
	\acro{LS}  {Link Specifications}
	\acro{LDIF}{Linked Data Integration Framework}
	\acro{LGD} {LinkedGeoData}
	\acro{LOD} {Linked Open Data}
	\acro{LOV} {Linked Open Vocabularies}
	\acro{LSTM}{Long Short-Term Memories}
	\acro{MSE}{Mean Squared Error}
	\acro{ML}{Machine Learning}
	\acro{MR}{Machine Learning}
	\acro{MRR}{Mean Reciprocal Rank}
	
	\acro{NIF}{Natural Language Processing Interchange Format}
	\acro{NIF4OGGD}{NLP Interchange Format for Open German Governmental Data}
	\acro{NLP}{Natural Language Processing}
	\acro{NER}{Named Entity Recognition}
	\acro{NMT}{Neural Machine Translation}
	\acro{NN}{Neural Network}
	\acrodefplural{NN}{Neural Networks}
	\acro{NLG}{Natural Language Generation}
	\acro{NED}{Named Entity Disambiguation}
	\acro{NERD}{Named Entity Recognition and Disambiguation}
	\acro{NL}{Natural Language}
	\acro{NIF}{NLP Interchange Format}
	\acro{NIF4OGGD}{NLP Interchange Format for Open German Governmental Data}
	\acro{NLP}{Natural Language Processing}
	\acro{NER}{Named Entity Recognition}
	\acro{NEL}{Named Entity Linking}
	\acro{NE}{Named Entity}
	\acro{NN}{Neural Network}
	\acro{NLI}{Natural Language Inference}
	\acro{OSM}{OpenStreetMap}
	\acro{OWL}{Web Ontology Language}
	\acro{OOV}{out-of-vocabulary}
	\acro{PFM}{Pseudo-F-Measures}
	\acro{PSO}{Particle-Swarm Optimization}
	\acro{PBSMT}{Phrase-Based Statistical Machine Translation}
	
	\acro{QA}{Question Answering}
	\acro{RDF}{Resource Description Framework}
	\acro{RBMT}{Rule-Based Machine Translation}
	\acro{RNN}{Recurrent Neural Network}
	\acro{ReLU}{rectified linear unit}
	\acro{RDFS}{RDF Schema}
	\acro{SKOS}{Simple Knowledge Organization System}
	\acro{SPARQL}{SPARQL Protocol and RDF Query Language}
	\acro{SRL}{Statistical Relational Learning}
	\acro{SWT}{Semantic Web Technologies}
	\acro{SW}{Semantic Web}
	\acro{SMT}{Statistical Machine Translation}
	\acro{SWMT}{Semantic Web Machine Translation}
	\acro{SKOS}{Simple Knowledge Organization System}
	\acro{SPARQL}{SPARQL Protocol and RDF Query Language}
	\acro{SRL}{Statistical Relational Learning}
	\acro{SF}{surface forms}

    \acro{TBMT} {Transfer-Based Machine Translation}
	\acro{UML}{Unified Modeling Language}
	\acro{USL}{Ukrainian Sign Language}
	\acro{URI}{Uniform Resource Identifier}
	\acro{WHO}{World Health Organization}
	\acro{WKT}{Well-Known Text}
	\acro{W3C}{World Wide Web Consortium}
	\acro{WSD}{Word Sense Disambiguation}
	\acro{WMT}{Workshop on Machine Translation}
    \acro{XML}{Extensible Markup Language}
	\acro{YPLL}{Years of Potential Life Lost}

	\acro{AOS}{Agricultural Ontology Services}
	\acro{AGRIS}{Agricultural Science and Technology}
	\acro{API}{Application Programming Interface}
	\acro{AOS}{Agricultural Ontology Services}
	\acro{AGRIS}{Agricultural Science and Technology}
	\acro{API}{Application Programming Interface}
	\acro{A2KB}{Annotation to Knowledge Base}
	\acro{BPSO}{Binary Particle-Swarm Optimization}
	\acro{BPMLOD}{Best Practices for Multilingual Linked Open Data}
	\acro{BPSO}{Binary Particle-Swarm Optimization}
	\acro{BPMLOD}{Best Practices for Multilingual Linked Open Data}
	\acro{BFS}{Breadth-First-Search}
	\acro{BPE}{Byte Pair Encoding}
	\acro{BoW}{Bag-of-Words}
	\acro{CBD}{Concise Bounded Description}
	\acro{COG}{Content Oriented Guidelines}
	\acro{CSV}{Comma-Separated Values}
	\acro{CBMT}{Corpus-Based Machine Translation}
	\acro{CLIR}{Cross-Language Information Retrieval}
	\acro{DPSO}{Deterministic Particle-Swarm Optimization}
	\acro{DALY}{Disability Adjusted Life Year}

	\acro{ER}{Entity Resolution}
	\acro{EM}{Expectation Maximization}
	\acro{EBMT}{Example-Based Machine Translation}
	\acro{EBNF}{Extended Backus--Naur Form}
	\acro{EL}{Entity Linking}
	\acro{FAO}{Food and Agriculture Organization of the United Nations}
	\acro{GIS}{Geographic Information Systems}
	\acro{GHO}{Global Health Observatory}
	\acro{GRU}{Gated recurrent unit}
	\acro{HDI}{Human Development Index}
	\acro{ICT}{Information and communication technologies}
	\acro{IFRS}{International Financial Reporting Standards}
	\acro{ICD}{International Classification of Diseases}
	\acro{IT}{Information Technology}
	\acro{IRI}{International Resource Identifier}
    \acro{KB}{Knowledge Base}
    \acro{KG}{Knowledge Graph}
    \acro{KGE}{Knowledge Graph Embeddings}
    \acro{KP}{Kronecker Product}
    \acro{KD}{Kronecker Decomposition}
	\acro{LR}  {Language Resource}
	\acro{LD}  {Linked Data}
	\acro{LLOD}  {Linguistic Linked Open Data}
	\acro{LIMES}{LInk discovery framework for MEtric Spaces}
	\acro{LS}  {Link Specifications}
	\acro{LDIF}{Linked Data Integration Framework}
	\acro{LGD} {LinkedGeoData}
	\acro{LOD} {Linked Open Data}
	\acro{LSTM}{Long Short-Term Memories}
	\acro{MSE}{Mean Squared Error}
	\acro{MWE}{Multiword Expressions}
	\acro{MT}{Machine Translation}
	\acro{ML}{Machine Learning}
	\acro{MR}{Machine Reading}
	\acro{MOS}{Manchester OWL Syntax}
	\acro{NIF}{Natural Language Processing Interchange Format}
	\acro{NIF4OGGD}{NLP Interchange Format for Open German Governmental Data}
	\acro{NLP}{Natural Language Processing}
	\acro{NER}{Named Entity Recognition}
	\acro{NMT}{Neural Machine Translation}
	\acro{NN}{Neural Network}
	\acro{NLG}{Natural Language Generation}
	\acro{NED}{Named Entity Disambiguation}
	\acro{NERD}{Named Entity Recognition and Disambiguation}
	\acro{NL}{Natural Language}
	\acro{NIF}{NLP Interchange Format}
	\acro{NIF4OGGD}{NLP Interchange Format for Open German Governmental Data}
	\acro{NLP}{Natural Language Processing}
	\acro{NER}{Named Entity Recognition}
	\acro{NEL}{Named Entity Linking}
	\acro{NE}{Named Entity}
	\acro{NN}{Neural Network}
	\acro{NLI}{Natural Language Inference}
	\acro{OSM}{OpenStreetMap}
	\acro{OWL}{Web Ontology Language}
	\acro{OOV}{out-of-vocabulary}
	\acro{PFM}{Pseudo-F-Measures}
	\acro{PSO}{Particle-Swarm Optimization}
	\acro{PBSMT}{Phrase-Based Statistical Machine Translation}
	
	\acro{QA}{Question Answering}
	\acro{RDF}{Resource Description Framework}
	\acro{RBMT}{Rule-Based Machine Translation}
	\acro{RNN}{Recurrent Neural Network}
	\acro{ReLU}{rectified linear unit}
	\acro{SKOS}{Simple Knowledge Organization System}
	\acro{SPARQL}{SPARQL Protocol and RDF Query Language}
	\acro{SRL}{Statistical Relational Learning}
	\acro{SWT}{Semantic Web Technologies}
	\acro{SW}{Semantic Web}
	\acro{SMT}{Statistical Machine Translation}
	\acro{SWMT}{Semantic Web Machine Translation}
	\acro{SKOS}{Simple Knowledge Organization System}
	\acro{SPARQL}{SPARQL Protocol and RDF Query Language}
	\acro{SRL}{Statistical Relational Learning}
	\acro{SF}{surface forms}
	\acro{SVM}{Support Vector Machines}

    \acro{TBMT} {Transfer-Based Machine Translation}
	\acro{UML}{Unified Modeling Language}
	\acro{USL}{Ukrainian Sign Language}
	\acro{URI}{Uniform Resource Identifier}
	\acro{WHO}{World Health Organization}
	\acro{WKT}{Well-Known Text}
	\acro{W3C}{World Wide Web Consortium}
	\acro{WSD}{Word Sense Disambiguation}
	\acro{WWW}{World Wide Web}
    \acro{XML}{Extensible Markup Language}
	\acro{YPLL}{Years of Potential Life Lost}

	\acro{AOS}{Agricultural Ontology Services}
	\acro{AGRIS}{Agricultural Science and Technology}
	\acro{API}{Application Programming Interface}
	\acro{BPSO}{Binary Particle-Swarm Optimization}
	\acro{BPMLOD}{Best Practices for Multilingual Linked Open Data}
	\acro{CBD}{Concise Bounded Description}
	\acro{COG}{Content Oriented Guidelines}
	\acro{CSV}{Comma-Separated Values}
	\acro{CBMT}{Corpus-Based Machine Translation}
	\acro{CLIR}{Cross-Language Information Retrieval}
	\acro{DPSO}{Deterministic Particle-Swarm Optimization}
	\acro{DALY}{Disability Adjusted Life Year}
	\acro{DRL}{Deep Reinforcement Learning}

	\acro{ER}{Entity Resolution}
	\acro{EM}{Expectation Maximization}
	\acro{EBMT}{Example-Based Machine Translation}
	\acro{EBNF}{Extended Backus--Naur Form}
	\acro{EL}{Entity Linking}
	\acro{FAO}{Food and Agriculture Organization of the United Nations}
	\acro{GIS}{Geographic Information Systems}
	\acro{GHO}{Global Health Observatory}
	\acro{HDI}{Human Development Index}
	\acro{ICT}{Information and communication technologies}
    \acro{KB}{Knowledge Base}
    \acro{KBSE}{Knowledge Base Semantic Embedding}
	\acro{LR}  {Language Resource}
	\acro{LD}  {Linked Data}
	\acro{LLOD}  {Linguistic Linked Open Data}
	\acro{LIMES}{LInk discovery framework for MEtric Spaces}
	\acro{LS}  {Link Specifications}
	\acro{LDIF}{Linked Data Integration Framework}
	\acro{LGD} {LinkedGeoData}
	\acro{LOD} {Linked Open Data}
	\acro{ML}{Machine Learning}
	\acro{MDP}{Markov Decision Process}
	\acro{MT}{Machine Translation}
	\acro{NIF}{Natural Language Processing Interchange Format}
	\acro{NIF4OGGD}{NLP Interchange Format for Open German Governmental Data}
	\acro{NLP}{Natural Language Processing}
	\acro{NER}{Named Entity Recognition}
	\acro{NMT}{Neural Machine Translation}
	\acro{NN}{Neural Network}
	\acro{NLG}{Natural Language Generation}
	\acro{NED}{Named Entity Disambiguation}
	\acro{NERD}{Named Entity Recognition and Disambiguation}
	\acro{NL}{Natural Language}
	\acro{OWL}{Web Ontology Language}
	\acro{QA}{Question Answering}
	\acro{RDF}{Resource Description Framework}
	\acro{RBMT}{Ru\-le-Ba\-sed Ma\-chi\-ne Trans\-la\-tion}
	\acro{REG}{Referring Expression Generation}
	\acro{SKOS}{Simple Knowledge Organization System}
	\acro{SPARQL}{SPARQL Protocol and RDF Query Language}
	\acro{SRL}{Statistical Relational Learning}
	\acro{SWT}{Semantic Web Technologies}
	\acro{SW}{Semantic Web}
	\acro{SMT}{Statistical Machine Translation}
	\acro{SWMT}{Semantic Web Machine Translation}

    \acro{TBMT} {Transfer-Based Machine Translation}
	\acro{VS}{Version Space}

	\acro{WHO}{World Health Organization}
	\acro{WKT}{Well-Known Text}
	\acro{W3C}{World Wide Web Consortium}
	\acro{WSD}{Word Sense Disambiguation}
    \acro{XML}{Extensible Markup Language}
	\acro{YPLL}{Years of Potential Life Lost}

	\acro{AOS}{Agricultural Ontology Services}
	\acro{AGRIS}{Agricultural Science and Technology}
	\acro{API}{Application Programming Interface}
	\acro{AOS}{Agricultural Ontology Services}
	\acro{AGRIS}{Agricultural Science and Technology}
	\acro{API}{Application Programming Interface}
	\acro{A2KB}{Annotation to Knowledge Base}
	\acro{BPSO}{Binary Particle-Swarm Optimization}
	\acro{BPMLOD}{Best Practices for Multilingual Linked Open Data}
	\acro{BPSO}{Binary Particle-Swarm Optimization}
	\acro{BPMLOD}{Best Practices for Multilingual Linked Open Data}
	\acro{BFS}{Breadth-First-Search}
	\acro{BPE}{Byte Pair Encoding}
	\acro{BoW}{Bag-of-Words}
	\acro{CBD}{Concise Bounded Description}
	\acro{COG}{Content Oriented Guidelines}
	\acro{CSV}{Comma-Separated Values}
	\acro{CBMT}{Corpus-Based Machine Translation}
	\acro{CLIR}{Cross-Language Information Retrieval}
	\acro{DPSO}{Deterministic Particle-Swarm Optimization}
	\acro{DALY}{Disability Adjusted Life Year}

	\acro{ER}{Entity Resolution}
	\acro{EM}{Expectation Maximization}
	\acro{EBMT}{Example-Based Machine Translation}
	\acro{EBNF}{Extended Backus--Naur Form}
	\acro{EL}{Entity Linking}
	\acro{FAO}{Food and Agriculture Organization of the United Nations}
	\acro{GIS}{Geographic Information Systems}
	\acro{GHO}{Global Health Observatory}
	\acro{GRU}{Gated recurrent unit}
	\acro{HDI}{Human Development Index}
	\acro{ICT}{Information and communication technologies}
	\acro{IFRS}{International Financial Reporting Standards}
	\acro{ICD}{International Classification of Diseases}
	\acro{IT}{Information Technology}
    \acro{KB}{Knowledge Base}
    \acro{KG}{Knowledge Graph}
    \acro{KGE}{Knowledge Graph Embedding}
	\acro{LR}  {Language Resource}
	\acro{LD}  {Linked Data}
	\acro{LLOD}  {Linguistic Linked Open Data}
	\acro{LIMES}{LInk discovery framework for MEtric Spaces}
	\acro{LS}  {Link Specifications}
	\acro{LDIF}{Linked Data Integration Framework}
	\acro{LGD} {LinkedGeoData}
	\acro{LOD} {Linked Open Data}
	\acro{LSTM}{Long Short-Term Memories}
	\acro{MSE}{Mean Squared Error}
	\acro{MWE}{Multiword Expressions}
	\acro{MT}{Machine Translation}
	\acro{ML}{Machine Learning}
	\acro{MR}{Machine Reading}
	\acro{NIF}{Natural Language Processing Interchange Format}
	\acro{NIF4OGGD}{NLP Interchange Format for Open German Governmental Data}
	\acro{NLP}{Natural Language Processing}
	\acro{NER}{Named Entity Recognition}
	\acro{NMT}{Neural Machine Translation}
	\acro{NN}{Neural Network}
	\acro{NLG}{Natural Language Generation}
	\acro{NED}{Named Entity Disambiguation}
	\acro{NERD}{Named Entity Recognition and Disambiguation}
	\acro{NL}{Natural Language}
	\acro{NIF}{NLP Interchange Format}
	\acro{NIF4OGGD}{NLP Interchange Format for Open German Governmental Data}
	\acro{NLP}{Natural Language Processing}
	\acro{NER}{Named Entity Recognition}
	\acro{NEL}{Named Entity Linking}
	\acro{NE}{Named Entity}
	\acro{NN}{Neural Network}
	\acro{NLI}{Natural Language Inference}
	\acro{OSM}{OpenStreetMap}
	\acro{OWL}{Web Ontology Language}
	\acro{OOV}{out-of-vocabulary}
	\acro{PFM}{Pseudo-F-Measures}
	\acro{PSO}{Particle-Swarm Optimization}
	\acro{PBSMT}{Phrase-Based Statistical Machine Translation}
	
	\acro{QA}{Question Answering}
	\acro{RL}{Reinforcement Learning}
	\acro{RDF}{Resource Description Framework}
	\acro{RNN}{Recurrent Neural Network}
	\acro{ReLU}{rectified linear unit}
	
	\acro{SKOS}{Simple Knowledge Organization System}
	\acro{SWT}{Semantic Web Technologies}
	\acro{SW}{Semantic Web}
	\acro{SPARQL}{SPARQL Protocol and RDF Query Language}
	\acro{SRL}{Statistical Relational Learning}
	\acro{SVM}{Support Vector Machines}
    \acro{SGD}{Stochastic Gradient Descent}

    \acro{TBMT} {Transfer-Based Machine Translation}
	\acro{UML}{Unified Modeling Language}
	\acro{USL}{Ukrainian Sign Language}
	\acro{URI}{Uniform Resource Identifier}
	\acro{WHO}{World Health Organization}
	\acro{WKT}{Well-Known Text}
	\acro{W3C}{World Wide Web Consortium}
	\acro{WSD}{Word Sense Disambiguation}
    \acro{XML}{Extensible Markup Language}
	\acro{YPLL}{Years of Potential Life Lost}
\end{acronym}  
\newcommand{\RealNumbers}{\ensuremath{\mathbb{R}}\xspace}

\newcommand{\kg}{\ensuremath{\mathcal{G}}\xspace}
\newcommand{\emb}[1]{\ensuremath{\mathbf{#1}}}

\newcommand{\triple}[3]{(\texttt{#1}, \texttt{#2}, \texttt{#3})}
\newcommand{\invtriple}[3]{(\texttt{#1}, $\texttt{#2}^{-1}$, \texttt{#3})}

\newcommand{\pair}[2]{(\texttt{#1}, \texttt{#2})}

\newcommand{\entities}{\ensuremath{\mathcal{E}}\xspace}
\newcommand{\relations}{\ensuremath{\mathcal{R}}\xspace}

\newcommand{\scoreFunc}{\phi}

\newcommand{\phat}{\hat{p}}
\newcommand{\headt}{\texttt{h}}
\newcommand{\relt}{\texttt{r}}
\newcommand{\tailt}{\texttt{t}}

\begin{document}
\title{Kronecker Decomposition for Knowledge Graph Embeddings}

\author{Caglar Demir}
\email{caglar.demir@upb.de}
\orcid{0000-0001-8970-3850}
\affiliation{%
  \institution{Data Science Group, Paderborn University}
  \city{Paderborn}
  \country{Germany}
}
\author{Julian Lienen}
\email{julian.lienen@upb.de}
\orcid{0000-0003-2162-8107}
\affiliation{%
  \institution{Heinz Nixdorf Institute, Paderborn University}
  \city{Paderborn}
  \country{Germany}
}
\author{Axel-Cyrille Ngonga Ngomo}
\email{axel.ngonga@upb.de}
\orcid{0000-0001-7112-3516}
\affiliation{%
  \institution{Data Science Group, Paderborn University}
  \city{Paderborn}
  \country{Germany}
}
\renewcommand{\shortauthors}{Demir et al.}

\begin{abstract}
Knowledge graph embedding research has mainly focused on learning continuous representations of entities and relations tailored towards the link prediction problem. 
Recent results indicate an ever increasing predictive ability of current approaches on benchmark datasets. 
However, this effectiveness often comes with the cost of over-parameterization and increased computationally complexity. 
The former induces extensive hyperparameter optimization to mitigate malicious overfitting. 
The latter magnifies the importance of winning the hardware lottery. 
Here, we investigate a remedy for the first problem. 
We propose a technique based on Kronecker decomposition to reduce the number of parameters in a knowledge graph embedding model, while retaining its expressiveness.
Through Kronecker decomposition, large embedding matrices are split into smaller embedding matrices during the training process.
Hence, embeddings of knowledge graphs are not plainly retrieved but reconstructed on the fly.
The decomposition ensures that elementwise interactions between three embedding vectors are extended with interactions within each embedding vector.
This implicitly reduces redundancy in embedding vectors and encourages feature reuse.
To quantify the impact of applying Kronecker decomposition on embedding matrices, we conduct a series of experiments on benchmark datasets. 
Our experiments suggest that applying Kronecker decomposition on embedding matrices leads to an improved parameter efficiency on all benchmark datasets. 
Moreover, empirical evidence suggests that reconstructed embeddings entail robustness against noise in the input knowledge graph. 
To foster reproducible research, we provide an open-source implementation of our approach, including training and evaluation scripts as well as pre-trained models.\footnote{\raggedright\url{https://github.com/dice-group/dice-embeddings}}
\end{abstract}

\maketitle

\section{Introduction}
\label{sec:intro}
~\ac{KGE} models learn continuous vector representations of entities and relations~\cite{nickel2015review,wang2017knowledge,trouillon2016complex,dettmers2018convolutional,sun2019rotate,demir2021convolutional}. 
These representations have been successfully applied in a wide range of applications including question answering, link prediction, and recommender systems~\citep{cai2018comprehensive,ji2020survey,eder2012knowledge,nickel2015review}. 

Most~\ac{KGE} models are designed to retain a linear complexity in the number of trainable parameters and in the number of triples to scale to large~\acp{KG}~\cite{bordes2013translating,yang2015embedding,trouillon2016complex,sun2019rotate,demir2021convolutional}. 
Although the number of parameters of most \ac{KGE} models grows linearly in the number of unique entities and relations, training on large \acp{KG} become a computationally challenging task. 
For example, when embedding Freebase with an embedding size of 200, DistMult requires 33 GB of main memory solely to store its parameters~\cite{dettmers2018convolutional}.
Such considerable memory requirement is still a barrier for efficient training and inference of embedding models~\cite{edalati2021kronecker}. 
Although there have been numerous attempts to reduce high memory usage
by means of decomposing pre-trained overparameterized language models (see DistilBERT~\cite{sanh2019distilbert}, TinyBERT~\cite{jiao2019tinybert}, MobileBERT~\cite{sun2020mobilebert}, ALP-KD\cite{passban2020alp}, MATE-KD~\cite{rashid2021mate}, and KnGPT2~\cite{edalati2021kronecker}), the decomposition of \ac{KGE} models have not yet been studied. 

In this paper, we investigate the problem of learning \emph{compressed} \acp{KGE}. 
Our aim is to reduce the number of explicitly stored parameters, while retaining the expressiveness of \ac{KGE} models. 
~\citet{trouillon2016complex} showed that finding the best ratio between expressiveness and the number of parameters is a keystone in successful applications of \ac{KGE} models. 
To increase the expressiveness of most \ac{KGE} models, increasing the embedding vector size is often the only option~\cite{dettmers2018convolutional}. 
Yet, solely increasing the parameters of models (a) makes models more prone to overfitting and (b) increases the hardware demand, i.e., the importance of winning the hardware lottery is magnified~\cite{hooker2021hardware}.
~\ac{KGE} literature shows that considerable amount of effort is often invested into hyperparameter optimization to find a good ratio between expressiveness and the number of parameters to mitigate
overfitting~\cite{zhang2019quaternion,sun2019rotate,ruffinelli2019you}.
With this consideration, we propose a technique based on \ac{KD}. 
Applying \ac{KD} on embedding matrices implicitly reduces redundancy in embedding vectors and encourages feature reuse (see Section~\ref{sec:methodology}). 
Through the decomposition, embeddings are not merely retrieved but reconstructed on the fly. 
This inherently reduces the number of explicitly stored parameters, and hence decreases the memory footprint of the decomposed models.
Importantly, our technique can be readily used in combination with many \ac{KGE} models.

To evaluate our technique, we conducted a series of experiments on benchmark datasets. 
On each benchmark dataset, we applied a fine-grained parameter sweep in the size of embedding vectors.
All competing \ac{KGE} models are trained via a standard training technique as in the literature (see Section~\ref{subsec:kge_training}).
We developed a hardware-agnostic embedding framework to conduct our experiments with an ease (see Section~\ref{subsec:implementation_details}).
To ensure the reproducibility of our experiments, we fixed the seed for the pseudo-random generator to 1. 
Overall, our results suggest that learning compressed embeddings via \ac{KD} reduces the number of parameters on each dataset with each hyperparameter configuration, while yielding competitive performance in the link prediction task (see Section~\ref{sec:results}). 
First series of our experiments indicated that \textbf{59\% parameter reduction} can be achieved via applying \ac{KD} on relation embeddings, while retraining the training and testing performance. 
Moreover, applying \ac{KD} on relation embeddings and entity embeddings results in on average \textbf{16.4 times fewer parameters}. 
We also evaluate link prediction performances of models with model calibration and under noisy input triples, since most predictions in the link prediction task are often uncalibrated and publicly available datasets involve noisy triples (see Section~\ref{subsec:model_calibration}).
Results of the former experiments suggest that MRR performance of all models are improved up to absolute $10\%$ provided that they suffer from overfitting. 
As the degree of overfitting decreases, model calibration techniques (see Label Smoothing and Label Relaxation in Section \ref{sec:background}) do not improve generalization performances. 
Results of the latter experiments suggest that learning compressed knowledge graph embeddings makes models more robust against noise in \acp{KG}. 
\section{Related Work}
\label{sec:related-work}
In the last decade, a plethora of \ac{KGE} approaches have been successfully applied to tackle various tasks~\citep{cai2018comprehensive,ji2020survey,nickel2015review,gligorijevic2021structure}. Here, we give a brief overview of selected \ac{KGE} approaches.

\citet{nickel2011three} proposed a three-way factorization of a third-order binary tensor representing the input \ac{KG}. The proposed approach (RESCAL) is limited in its scalability as it has a quadratic complexity in the factorization rank~\cite{nickel2016holographic}.
~\citet{yang2015embedding} proposed DistMult that can be seen as an extension of RESCAL with a diagonal matrix per relation, where the dense core tensor is replaced with diagonal matrices. 
DistMult does not perform well on triples with antisymmetric relations (e.g.~\triple{Barack}{HasChild}{Malia}), whereas it performs well on symmetric relations (e.g.~\triple{Barack}{Married}{Michelle}). 
To avoid this shortcoming,~\citet{trouillon2016complex} extended DistMult by learning representations in a complex vector space.
Their approach (ComplEx) is able to infer both symmetric and antisymmetric relations via a Hermitian inner product of embeddings which involves the conjugate-transpose of one of the two input vectors. 
Motivated by learning complex-valued embeddings,~\citet{sun2019rotate} designed RotatE which employs a rotational model taking predicates as rotations from subjects to objects in complex space via the element-wise Hadamard product. 
QuatE extends ComplEx into quaternions through applying the quaternion multiplication followed by an inner product to compute scores of triples~\citep{zhang2019quaternion}.

All aforementioned approaches learn embeddings of entities and relations through capturing multiplicative based interactions. 
Although such approaches perform well in terms of predictive accuracy and computational complexity, to increase their expressiveness the embedding vector size is the only option.~\citet{dettmers2018convolutional}, 
~\citet{nguyen2017novel}, and ~\citet{balavzevic2019hypernetwork} have shown that convolution operation can be applied to increase the expresiveness of \ac{KGE} models without significantly increasing the number of parameters. 
ConvE applies a 2D convolution operation to model the interactions between entities and relations~\citep{dettmers2018convolutional}. 
ConvKB extends ConvE by omitting the reshaping operation in the encoding of representations in the convolution operation~\citep{nguyen2017novel}. 
Similarly, HypER extends ConvE by applying relation-specific 1D convolutions as opposed to applying filters from concatenated subject and relation vectors~\citep{balavzevic2019hypernetwork}.~\citet{demir2021convolutional,demir2021hyperconvolutional} extended ConvE through combining 2D convolution operation with Hermitian inner product, Quaternion and Octonion multiplications.
%
\section{Background}
\label{sec:background}

\subsection{Knowledge Graphs \& Link Prediction}
\label{subsec:kg_lp}
Let \entities\ and \relations\ denote a set of entities and relations. A knowledge graph (\ac{KG}) can be defined as a set of triples $\kg  \subseteq \entities \times \relations \times \entities$, where each triple $\triple{h}{r}{t} \in \kg$ contains two entities $\texttt{h},\texttt{t} \in \entities$ and a relation $\texttt{r} \in \relations$~\cite{demir2021convolutional}. Hence, knowledge graphs represent structured collections of facts in the form of typed relationships between entities~\citep{hogan2020knowledge}. These collections of facts have been used in a wide range of applications, including web search, question answering, and recommender systems~\citep{nickel2015review}. Yet, most knowledge graphs on the web are far from complete. The link prediction task on \acp{KG} refers to predicting whether a triple is likely to be true~\cite{dettmers2018convolutional,demir2021convolutional}. 
This task is often formulated as the problem of learning a parametrized scoring function $\scoreFunc_\Theta: \entities \times \relations \times \entities \rightarrow \mathbb{R}$ such that $\scoreFunc_\Theta\triple{h}{r}{t}$ ideally signals the likelihood of \triple{h}{r}{t} is true~\cite{ji2020survey}. 
Here, $\Theta$ contains embeddings of entities and relations along with other trainable parameters. 
For instance, given the triples \triple{Barack}{Married}{Michelle} and $\triple{Michelle}{HasChild}{Malia}\in \kg$, a good scoring function is expected to return high scores for \triple{Barack}{HasChild}{Malia} and
\triple{Michelle}{Married}{Barack}, while returning a considerably lower score for \triple{Malia}{HasChild}{Barack}.

\subsection{Knowledge Graph Embeddings and Training Strategies}
\label{subsec:kge_training}
Most knowledge graph embedding (\ac{KGE}) models are designed to learn continuous vector representations of entities and relations tailored towards predicting missing triples. 
In our notation, the embedding vector of the entity $e \in \entities$ is denoted by $\mathbf{e} \in \mathbb{R}^{d_e}$ and the embedding vector for the relation $r \in \relations$ is denoted by $\mathbf{r} \in \mathbb{R}^{d_r}$. 
Embedding vectors are learned through minimizing a loss function (see Equation~\ref{eq:ce_loss}) with a first-order optimizer.
The resulting models are then evaluated w.r.t. their ability of predicting missing entity rankings~\cite{ruffinelli2019you}. 

Three training strategies are commonly used for \ac{KGE} models.~\citet{bordes2013translating} designed a negative sampling technique via perturbing an entity in a randomly sampled triple. 
In this setting, a triple $\triple{h}{r}{t} \in \kg$ is considered as a positive example, whilst 
$ \{ \triple{h}{r}{x} \mid \forall x \in \entities\} \cup \{ \triple{x}{r}{t} \mid \forall x \in \entities \}$ is considered as a set of possible candidate negative examples. 
For each positive triple $\triple{h}{r}{t} \in \kg$, a negative triple is sampled from the set of corresponding candidate negative triples.
~\citet{kotnis2017analysis} analysed the impact of variations of the negative sampling technique designed by~\citet{bordes2013translating} on the link prediction task.
~\citet{lacroix2018canonical} discarded the idea of randomly sampling negative triples and proposed 1vsAll/1vsN the training strategy. 
For each positive triple $\triple{h}{r}{t} \in \kg$, all possible tail perturbed set of triples are considered as negative triples regardless of whether a perturbed triple exists in the input knowledge graph \ac{KG} ($\{ \triple{h}{r}{x} | \forall x \in \entities: x \not= t \}$) as similarly done by~\citet{bordes2013translating}. 
Given that this setting does not involve negative triples via head perturbed entities, a data augmentation technique is applied to add inverse triples (also known as reciprocal triples~\cite{balavzevic2019tucker}) \invtriple{t}{r}{h} for each \triple{h}{r}{t}. 
Their results showed that even simpler models reached state-of-the-art performance in link prediction task via 1vsAll. 
This partially stems from the fact that the ability of modelling antisymmetric relations is not required due to learning inverse relation representations. 
In the 1vsAll training strategy, a training data point consists of \pair{h}{r} and a binary vector containing a single "1" for the \texttt{t} and "0"s for other entities.
~\citet{dettmers2018convolutional} extended 1vsAll into KvsAll\footnote{We use the terminology introduced by~\citet{ruffinelli2019you}.} via constructing multi-label binary vectors for each \pair{h}{r}. 
More specifically, a training data point consists of a pair \pair{h}{r} and a binary vector containing  "1" for $\{x | x\in \entities \wedge \triple{h}{r}{x} \in \kg \}$ 
and "0"s for other entities. 
During training, for a given pair \pair{h}{r}, predicted scores (logits) for all entities are computed, i.e.,
$ \forall x \in \entities: \scoreFunc(\triple{h}{r}{x})) =: \mathbf{z} \in \mathbb{R}^ {|\entities|}$. Through 
the logistic sigmoid function $\sigma(\mathbf{z}) = \frac{1}{1 + \text{exp}(-\mathbf{z})}$, scores are normalized to obtain predicted probabilities of entities denoted by $\mathbf{\hat{y}}$ . A loss incurred on a training data point is then computed as 
\begin{equation}
l(\mathbf{\hat{y}},\mathbf{y}) = -\frac{1}{|\entities|}\sum\limits_{i=1}^{|\entities|}  \mathbf{y}^{(i)} \text{log}(\hat{\mathbf{y}}^{(i)} + \big( 1-\mathbf{y}^{(i)} \big) \text{log}\big( 1-\hat{\mathbf{y}}^{(i)} \big),
\label{eq:loss}
\end{equation}
where $\mathbf{y} \in [0,1]^{|\entities|}$ is the binary sparse label vector. If $(\texttt{h},\texttt{r},\texttt{e}_i ) \in \kg$, then  $\mathbf{y}^{(i)} = 1$, otherwise $\mathbf{y}^{(i)} = 0$. 
Recent works show that learning $\Theta$ by means of minimizing Equation~\ref{eq:loss} often leads to state-of-the-art link prediction performance~\cite{balavzevic2019tucker,balavzevic2019hypernetwork,dettmers2018convolutional,demir2021convolutional}.
Expectedly, 1vsAll and Kvsall are computationally more expensive than the negative sampling. 
As $|\entities|$ increases, 1vsAll and KvsAll training strategies  become less applicable. Yet, recent \ac{KGE} models are commonly trained with 1vsAll or KvsAll~\cite{ruffinelli2019you}.


%
\subsection{Model Calibration and Robustness}
\label{subsec:model_calibration}

Section~\ref{subsec:kge_training} elucidated that most \ac{KGE} models are trained to predict missing entities in a fashion akin to multi-class (1vsAll) and multi-label (KvsAll) classification problems.
\citet{tabacof2019probability} showed that previous state-of-the-art \ac{KGE} models generate uncalibrated predictions, i.e.,
probability estimates assigned to triples are unreliable. In many recent works, a model calibration technique has been applied to obtain calibrated predictions~\cite{dettmers2018convolutional,balavzevic2019hypernetwork,balavzevic2019tucker,demir2021convolutional,demir2021hyperconvolutional}. 

A \ac{KGE} model trained via 1vsAll or Kvsall can be interpreted as a probabilistic mapping $\phat : \entities \times \relations \mapsto \mathbb{P}(\mathcal{Y})^{|\entities|}$ with $\mathcal{Y}=\{0,1\}$ representing the events that the respective entities apply ($1$) or not ($0$). 
Hence, the training information is in the simplest form rendered as probabilities $p(\texttt{t} \, | \, \texttt{h}, \texttt{r}) := p( \mathbbm{1}(\texttt{t} \in \relt(\headt)) \in \mathcal{Y} \, | \, \texttt{h}, \texttt{r}) = 1 \, \text{s.t.} \, \forall \triple{h}{r}{t} \in \mathcal{G}$, e.g., the probability of the triple $\triple{Michelle}{HasChild}{Malia}\in \kg$ would be set to $1$. 
Then, Equation~\ref{eq:loss} can be rewritten as 
\begin{equation}
    L = -\frac{1}{|\entities|} \sum_{\tailt \in \entities} p(\tailt \, | \, \headt, \relt) \log \phat(\tailt \, | \, \headt, \relt) + (1 - p(\tailt \, | \, \headt, \relt) ) \log (1 - \phat(\tailt \, | \, \headt, \relt)).
    \label{eq:ce_loss}
\end{equation}
The calibration of $\phat$ characterizes the convergence between the model confidence and actual accuracy, ideally matching each other~\cite{guo2017calibration}. 
Not only can overconfident but incorrect predictions result into harmful consequences, also too modest predictions can leave much potential unused. 
Hence, well-calibration can be seen as a second optimization goal in knowledge graph embedding learning. 
Training models with nondegenerate probability distributions \textemdash hard target values 1/0 \textemdash~provokes overconfidence in predictions as the learner is urged to reproduce extreme probabilities~\cite{tabacof2019probability}.
To weaken this issue, a plethora of calibration means has been proposed, commonly by leveraging explicit calibration data, such as isotonic regression \cite{isotonicregression}, Bayesian binning \cite{bayesianbinning} or temperature scaling \cite{guo2017calibration}. Besides calibration, model \textit{robustness} against label noise further plays an essential role in making a embedding model applicable to real-world data. Large knowledge graphs observed in practice typically comprise erroneous relations, misleading the learner when used as training instance~\cite{heindorf2017wsdm,heindorf2016vandalism,heindorf2019debiasing}. For instance, Albert Einstein has the type of \url{https://dbpedia.org/ontology/Eukaryote} that is defined a subclass of Fungus or \url{https://dbpedia.org/page/Boo_(dog)} has the type of Scientist and Dog. 




\subsubsection{Label Smoothing}

\citet{szegedy2016rethinking} designed the Label Smoothing technique (LS) that combines favorable properties in both quality respects discussed above. Not only can it be considered as an implicit calibration method \cite{DBLP:conf/nips/MullerKH19}, it has also shown improved robustness capabilities over the cross-entropy loss \cite{DBLP:conf/icml/LukasikBMK20}. 
LS transforms the probability distributions $p$ as defined above to less extreme distributions $p^s$ by distributing a certain amount of probability mass among all other events, in its simplest formulation uniformly. To this end, let $\alpha\in(0,1)$ be the smoothing parameter. 
Then, the smoothed training distributions are formally given by $p^s(\texttt{t} \, | \, \texttt{h}, \texttt{r}) = 1 - \alpha / 2 \; \forall \triple{h}{r}{t} \in \mathcal{G}$. 
These distributions $p^s$ replace $p$ in the cross-entropy loss formulation. Referring to the running example, label smoothing would assign $p^s(\texttt{Malia} \, | \, \texttt{Michelle}, \texttt{HasChild}) = 1 - \alpha / 2$ and $\alpha / 2$ to the complementary event.
By this, the model is no longer urged to reproduce probabilities approaching $1$, resulting in less over-confidence.
\subsubsection{Label Relaxation}
Albeit LS undeniably improves generalization performances, it can be questioned from a data modelling perspective. The smoothed distribution $p^s$, by default determined by a uniform smoothing policy, is unlikely to match the exact underlying ground-truth probability, introducing a modeling bias that may distort the learning in a less extreme yet still unrealistic way. Coming back to our previous example, the smoothed target probability $p^s(\mathbbm{1}(\texttt{Malia} \notin \texttt{HasChild}(\texttt{Michelle})) \, | \, \texttt{Michelle}, \texttt{HasChild}) = \alpha / 2$ seems to be implausible in case there is clear evidence for the truthfulness of the triple, and the learner would be incited to assign less probability on purpose.

As a generalization to mitigate such issues, so-called \textit{label relaxation} (LR) \cite{DBLP:conf/aaai/LienenH21} models the target as (credal) \textit{set} $\mathcal{Q}_\alpha \subseteq \mathbb{P}(\mathcal{Y})$ of candidate probability distributions ``sufficiently close'' to the original distribution $p$, whereby the exact ground-truth distribution is assumed to be in $\mathcal{Q}_\alpha$. To construct such target sets, LR assigns a certain amount of plausibility $\alpha \in (0,1]$ to other outcomes than the one observed in the training data. Transferred to KGE learning, the parameter $\alpha$ can be interpreted as an upper probability for the event that the respective entity does not apply to a pair $(\headt, \relt)$. Hence, the credal set $\mathcal{Q}^{\triple{h}{r}{t}}_\alpha$ of the triple $\triple{h}{r}{t} \in \mathcal{G}$ is given by
\begin{equation*}
    \mathcal{Q}^{\triple{h}{r}{t}}_\alpha := \{ p'(\cdot \, | \, \headt, \relt) \in \mathbb{P}(\mathcal{Y}) \, | \, p'(\tailt \, | \, \headt, \relt) \geq 1 - \alpha \}.
\end{equation*}
As a result, instead of committing to a particular distribution, the resulting target set $\mathcal{Q}_\alpha$ is less precisely but more reliably expressing the belief about the ground-truth probability, reducing the risk of entailing an undesirable bias in the target modelling. Relating to the ongoing example relation, the target probability $p$ is expected to be $p(\texttt{Malia} \, | \, \texttt{Michelle}, \texttt{HasChild}) \in [1-\alpha, 1]$, i.e., it must not necessarily be the smoothed nor the degenerate distribution. In the following, we will drop the dependence of the target set on $\triple{h}{r}{t}$, i.e., $\mathcal{Q}_\alpha := \mathcal{Q}^{\triple{h}{r}{t}}_\alpha$, for the sake of brevity.

To optimize a probabilistic model provided such (weak) supervision, the learner is then trained by comparing $\phat$ to the most plausible candidate $p' \in \mathcal{Q}_\alpha$ among the distributions in the target set in terms of the Kullback-Leibler divergence $D_{KL}(p || \phat) = \sum_{y \in \mathcal{Y}} p(y) \log \phat(y)$ based on the current model belief $\phat$. Leveraging the framework of \textit{optimistic superset learning} \cite{osl}, the LR loss is formally defined as 
\begin{equation*}
    L^*(\mathcal{Q}_\alpha, \hat{p}) = \min_{p' \in \mathcal{Q}_\alpha} D_{KL} \big( p' \mid \mid \hat{p} \big),
\end{equation*}
which simplifies to the closed-form
\begin{equation*}
    L^*(Q_\alpha , \hat{p}) = \begin{cases}
        0  & \text{if } \hat{p} \in Q_\alpha \\
        D_{KL}(p' || \hat{p}) & \text{otherwise} ,
    \end{cases}
\end{equation*}
where $p'(\mathbbm{1}(\tailt \in \relt(\headt)) \, | \, \headt, \relt) = 1 - \alpha$ and $\alpha$ otherwise. Roughly speaking, the consideration of $p'$ as closest, and thus most plausible, distribution to $\phat$ in the target set constitutes a form of optimism in the validity of $\phat$.
This loss instantiation has been proven to be convex and can be efficiently optimized. Results signal improved model calibration while retaining the generalization performance as achieved by LS \cite{DBLP:conf/aaai/LienenH21,Lienen2021_CSSL}. 
%
\subsection{Kronecker Product and Decomposition}
\label{subsec:kronecker_product}
For any matrix $\mathbf{X} \in \RealNumbers^{m \times n}$ and $\mathbf{Y} \in \RealNumbers^{m \times n}$, the Hadamard product $\mathbf{X} \circ \mathbf{Y}$ is defined as 
\begin{equation}
\mathbf{X} \circ \mathbf{Y} = \begin{bmatrix} 
    \mathbf{X}_{11}\mathbf{Y}_{11} & \dots  & \mathbf{X}_{1n}\mathbf{Y}_{1n} \\
    \vdots & \ddots & \vdots \\
    \mathbf{X}_{m1}\mathbf{Y}_{m1} & \dots  & \mathbf{X}_{mn}\mathbf{Y}_{mn} 
    \end{bmatrix} \in \RealNumbers^{m\times n},
\label{eq:hadamard_product}
\end{equation}
where $\mathbf{X}_{ij}$ \textbf{only interacts} with $\mathbf{Y}_{ij}$. For any matrix $\mathbf{Z} \in \RealNumbers^{p \times q}$, the \ac{KP}  $\mathbf{X} \otimes \mathbf{Z}$ is a block matrix:
\begin{equation}
\mathbf{X} \otimes \mathbf{Z} = \begin{bmatrix} 
    \mathbf{X}_{11}\mathbf{Z} & \dots  & \mathbf{X}_{1n}\mathbf{Z} \\
    \vdots & \ddots & \vdots \\
    \mathbf{X}_{m1}\mathbf{Z} & \dots  & \mathbf{X}_{mn}\mathbf{Z} 
    \end{bmatrix} \in \RealNumbers^{mp \times nq},
\label{eq:kronocker_product}
\end{equation}
where \textbf{every element} of $\mathbf{X}$ \textbf{interacts with every element} of $\mathbf{Z}$. 
In contrast to the Hadamard product, the Kronecker product is not commutative, i.e., $\mathbf{X} \otimes \mathbf{Z} \not = \mathbf{Z} \otimes \mathbf{X}$ most commonly holds. For more details, we refer to~\cite{van2000ubiquitous,graham2018kronecker}. Numerous works have shown that \ac{KP} defined in Equation~\ref{eq:kronocker_product} can be effectively applied to decompose a large matrix into two smaller matrices~\cite{greenewald2016robust,greenewald2013kronecker,cohen2019differential,tahaei2021kroneckerbert,zhang2021beyond}. 
In the Kronecker Decomposition (\ac{KD}), a large weight matrix of $\mathbf{W}\in R^{mp\times nq}$ can be decomposed into any $\mathbf{X} \in \mathbb{R}^{m_1  \times n_1} $ and $ \mathbf{Z} \in \mathbb{R}^{ \frac{mp}{m_1} \times \frac{nq}{n_1}}$. Different compression factors can be achieved through different shape configurations of smaller matrices. 
Tahaei et. al.~\cite{tahaei2021kroneckerbert} have shown that the following formulation can be effectively used to compute a linear transformation encoded in a weight matrix $\mathbf{W}$ via \ac{KD}: 
\begin{equation}
\Big( \mathbf{X}\otimes \mathbf{Z} \Big)x=\mathcal{V} \Big(\mathbf{Z} \; \mathcal{R}_{\frac{nq}{n_1} \times n_1}(x) \mathbf{X}^\top \Big), \label{eq:implicit_kronecker_decomposition}
\end{equation}
where $x \in \mathbb{R}^{nq}$ represent input feature vector, $\mathcal{V}: \mathbb{R} \to \mathbb{R}^{mp}$ flattens an input matrix into a vector, $ \mathcal{R}_{ \frac{nq}{n_1} \times n_1} $ converts x to a $\frac{nq}{n_1}$ by $n_1$ matrix by dividing the vector to columns of size $\frac{nq}{n_1}$ and concatenating the resulting columns together~\cite{tahaei2021kroneckerbert,lutkepohl1997handbook}. The computation defined in \ref{eq:implicit_kronecker_decomposition}
reduces the number of floating point operations required to perform \ac{KD} from $(2 m_1 m_2 -1) n_1 n2 $ to $min((2n_2 -1)m_2 n_1 + (2n_1 -1) m_2 m_1 ,(2 n_1 -1)n_2 m_1 + (2 n_2 -1) m_2 m_1 )$, where $n_1= \frac{mp}{m1}$ and $n_2= \frac{nq}{n_1}$~\cite{tahaei2021kroneckerbert}. In the next section, we elucidate our methodology on applying \ac{KD} in \ac{KGE}.
\section{Methodology}
\label{sec:methodology}
Previous works have shown that pre-trained overparameterized language models can be effectively decomposed into smaller weight matrices~\cite{sanh2019distilbert,jiao2019tinybert,sun2020mobilebert,rashid2021mate}.
Particularly, the Kronecker Decomposition (\ac{KD}) elucidated in Section~\ref{subsec:kronecker_product} leads a significant reduction in the number of parameters with at most a mild cost of predictive accuracy~\cite{edalati2021kronecker,tahaei2021kroneckerbert}. Analogous to the aforementioned two works, findings of~\citet{zhang2021beyond} and~\citet{wu2016compression} suggest that training a neural network via \ac{KD}
on weight matrices results in a significant parameter efficiency.
We are interested in learning
compressed \ac{KGE} by applying \ac{KD} during training. We aim to design a generic technique that can be applied in existing 
knowledge graph embedding (\ac{KGE}) models to reduce the number of explicitly stored parameters while retaining their expressiveness. 
Importantly, through \ac{KD} on embedding matrices, we aim to capture interactions within an embedding vector without requiring additional parameters. This is expected to encourage parameter reuse and reduces redundancy in model's parameters~\cite{huang2017densely}.

\subsection{Kronecker Decompression for Knowledge Graph Embeddings}
\label{subsec:kd_kge}
Most \ac{KGE} models are designed as a parametrized scoring function 
(say DistMult~\cite{yang2015embedding}) $\scoreFunc_\Theta: \entities \times \relations \times \entities \to \mathbb{R}$ that maps an input triple $\triple{h}{r}{t} \in \kg$ to a scalar value that is mapped to the unit interval using the sigmoid/logistic function. 
This normalized scalar value is interpreted to reflect the likelihood of \triple{h}{r}{t} is true which is denoted with $\phat(\texttt{t} \, | \, \texttt{h}, \texttt{r})$ in Section~\ref{subsec:model_calibration}. 
$\Theta$ denotes the parameters of $\scoreFunc$ that consists of an entity embedding matrix $\mathbf{E} \in \mathbb{R}^{|\entities| \times d}$, a relation embedding matrix $\mathbf{R} \in \mathbb{R}^{|\relations| \times d}$, and other trainable parameters, such as affine transformation over input embeddings, convolutions, batch normalization or instance normalization. 
Assume that $\scoreFunc=:\text{DistMult}$ and $\Theta:=\{ \mathbf{E},\mathbf{R} \}$, for a given $\triple{h}{r}{t} \in \kg$, a score is computed as 
\begin{equation}
    \text{DistMult}\triple{h}{r}{t}= \emb{h}\circ \emb{r} \cdot \emb{t}.
    \label{eq:distmult}
\end{equation}
In Equation~\ref{eq:distmult}, a triple score is computed thorough elementwise interactions between 3 $d$-dimensional real-valued embedding vectors, e.g., given $d=2$ and, $a,b,c,d,e,f \in \mathbb{R}$, a triple score is $[a \; b] \circ [c \; d] \cdot [e \; f] = ace + bdf$
. During training, the gradient of the loss function (see Equation \ref{eq:ce_loss}) w.r.t. 3 d-dimensional embedding vectors $\emb{h}, \emb{r}, \emb{t}$ is computed. 
The gradient of the loss w.r.t. the first item $a$ in $\emb{h}$ is computed in two steps:
the gradient of the loss. w.r.t. the prediction is computed and distributed over the addition operation and the elementwise multiplication via $ce$. 
Yet, the interaction between $(a,b)$ as well as
between $(a,a)$ are ignored, although $a$ and $b$ constituted $\emb{h}$ together. 
This stems from computing a scalar value via elementwise operations. 
These ignored interactions can be captured through applying \ac{KD} on embedding vectors as follows
\begin{equation}
  \begin{aligned}
  \; [aa \; ab \; ab \; bb] \circ [cc \; cd \; cd \; dd] \cdot [ee \; ef \; ef \; ff] & = aa cc e \\
               & + 2(ab cd ef) \\
               & + bb dd ff \; .
  \end{aligned}
 \label{eq:triple_score_via_blockwise_interaction}
\end{equation}
Through the middle term in Equation~\ref{eq:triple_score_via_blockwise_interaction}, interactions between $(a,b)$ and $(a,a)$ are incorporated without requiring additional parameters.
Hence, we argue that DistMult defined in Equation~\ref{eq:distmult} is less expressive than its \ac{KD} variation defined in Equation ~\ref{eq:triple_score_via_blockwise_interaction}. 
Consequently, if the both models are trained properly, the latter model is expected to perform at least as well as the former model in the link prediction problem. Importantly, the number of possible interactions within an embedding vector grows by $(d+1)^2$ when $d$ grows by $1$. 
Although these interactions are expected to encourage \emph{feature reuse} and \emph{reduce redundancy} in model's parameters, the computation of \ac{KP} on embedding vectors may become a bottleneck due to high computationally complexity of \ac{KD}. 
In this setting, $\mathbf{E}$ and $\mathbf{R}$ can be seen as compressed embeddings of entities and relations. 
Hence, given a triple \triple{h}{r}{t}, their compressed embeddings are retrieved and a triple score is computed via 
their decompressed embeddings. Our technique can be readily applied on many multiplicative based \ac{KGE} models. 
For the numerical stability, the batch normalisation or layer normalization can be applied to reduce the dependence of gradients on the scale of embedding vectors as these normalization techniques have beneficial effect on the gradient flow through models~\cite{ioffe2015batch,ba2016layer}.  
\subsection{Kronecker Decomposition for Relation Embeddings}
\label{subsubsec:kd_for_rel_emb}
\ac{KD} on relation embeddings of DistMult can be applied as 
\begin{equation}
    \text{\ac{KD}-Rel-DistMult}\triple{h}{r}{t}=\emb{h} \circ (\emb{r}\otimes\emb{r}) \cdot \emb{t},
    \label{eq:kd_rel_distMult}
\end{equation}
where $\emb{h}, \emb{t} \in \mathbf{E} : \mathbf{E} \in \mathbb{R}^d$, and $\emb{r} \in \mathbf{R}: \mathbf{R} \in \mathbb{R}^{\sqrt{d}}$. The parameter gain can be computed as $(d - \sqrt{d}) \times (|\relations|)$. Note that scores of triples are still computed based on 3 d-dimensional embedding vectors. As the size of the input graph and the number of relations grow, the parameter gain becomes more tangible.
%
\subsection{Kronecker Decomposition of Embeddings for 1vsAll}
\label{subsubsec:kd_with_linear_trans}
Most \acp{KG} contain more entities than relations (see Table~\ref{table:datasets}). 
Hence, the potential parameter gain of applying \ac{KD} on entity embeddings is expectedly larger than applying \ac{KD} on relation embeddings. 
Yet, applying \ac{KD} on tail entities embeddings in 1vsAll or KvsAll increase the runtime and memory requirements as these training strategies require considering all entities to compute an incurred loss for a single prediction (see Section~\ref{subsec:kge_training}). 
To alleviate this limitation, \ac{KD} can be applied as 
\begin{equation}
        \text{\ac{KD}-DistMult}\triple{h}{r}{t}= \; \sum_i ^{d} R( \big(\emb{h}\otimes\emb{h}) \circ (\emb{r}\otimes\emb{r}) \big)_i \cdot \emb{t}
        \label{eq:kd_head_rel_distmult},
\end{equation}
where $R(\cdot)$ reshapes a $d^2$ dimensional vector into $d$ by $d$ matrix.
The matrix vector product of the resulting vector and $\emb{t}$ is summed to obtain a score for an input triple \triple{h}{r}{t}.

In Section~\ref{subsubsec:kd_for_rel_emb} and \ref{subsubsec:kd_with_linear_trans}, we elucidated applying KD on embedding vectors. Yet, KD can be also applied to decompose linear transformation weight matrices in feed-forward and convolutional based embedding models during training.
\section{Experimental Setup}
\subsection{Training and Optimization}
\label{subsec:training_and_optimization}
We trained approaches with the 1vsAll training strategy (see Section~\ref{subsec:kge_training}) as commonly done in the literature~\cite{lacroix2018canonical,ruffinelli2019you}. 
All models are trained with the ADAM optimizer~\citep{kingma2014adam} and minimize the cross entropy loss function (see Equation~\ref{eq:ce_loss}). 
We use the same loss function for all approaches on all datasets as~\citet{mohamed2019loss} previously showed that generalization performance of \ac{KGE} models can be significantly influenced by the choice of the loss function. Moreover, we apply the batch normalization~\cite{ioffe2015batch} to facilitate numerical stability and accelerate convergence during training. We trained the model for $1000$ epochs with a learning rate of $.01$ and a batch size of $1024$.
We further optimized the embedding vector sizes in $\{4,9,16,25,36,49,64,81,100,121,144,169,196,225,289,324,361,400\}$ using a grid search. For label smoothing and label relaxation, we consider $\alpha \in \{ .1,.2 \}$. This results in training models with soften target values, i.e., $0 <\textbf{y}^{(i)} < 1$. 
Through large parameter sweep in the embedding vector size, we aim to obtain a fine-grained performance analysis. 
As elucidated in Section~\ref{sec:methodology}, we hypothesize that the impact of learning compressed embeddings becomes more tangible as the size of the embedding vector increases. 
Importantly, we also share test and train performance of all operations with all configurations. By doing so, we aim to quantify the impact of applying label smoothing and label relaxation during training and testing separately. 
We do not apply an explicit regularization (e.g.  L1, L2 regularization or the Dropout technique~\cite{srivastava2014dropout}) as regularization techniques may not allow to observe any regularization effect of label smoothing and label relaxation.
\subsection{Datasets}
\label{subsec:dataset}
We evaluated our proposed models on standard link prediction benchmark datasets (KINSHIP and UMLS datasets)~\cite{trouillon2017complex}. The Kinships knowledge graph describes the 26 different kinship relations of the Alyawarra tribe and the unified medical language system (UMLS) knowledge graph describes 135 medical entities via 49 relations describing~\citep{trouillon2017complex}. Note that we omitted WN18RR, FB15K-237, and YAGO3-10 from our experiments for two reasons: First, 
we aim to conduct experiments to quantify the impact of Kronecker Decomposition in a fine-grained many unique configurations with very large number of epochs (1000). 
KINSHIP and UMLS datasets are considerably smaller datasets compared to WN18RR, FB15K-237, and YAGO3-10. Hence, training three models with 21 unique configurations for 1000 epochs  (see Section~\ref{subsec:training_and_optimization}) on WN18RR, FB15K-237, and YAGO3-10 can take up to a month. 
Moreover, two recent works showed that these datasets involve entities on the validation and test data splits that do not occur in the train split~\cite{libkge,demir2021out}.
\begin{table}
    \caption{Overview of datasets.}
    \centering
    \begin{tabular}{l c c c c c}
    \toprule
    \bfseries Dataset & \multicolumn{1}{c}{$|\entities|$}&  $|\relations|$ & $|\kg^{\text{Train}}|$ & $|\kg^{\text{Val.}}|$ &  $|\kg^{\text{Test}}|$\\
    \midrule
    UMLS             &  136    &     93  &    10,432  &  1304 &  1965 \\
    KINSHIP          &  105    &     51  &    17,088  &  2136 &  3210 \\
    \bottomrule
    \end{tabular}
    \label{table:datasets}
\end{table}

\subsection{Evaluation metrics and Baseline Selection}
\label{subsec:evaluation_metrics}
We use the standard metrics \textit{filtered} \ac{MRR} and hits at N (H@N) for link prediction~\cite{dettmers2018convolutional,balavzevic2019tucker,balavzevic2019hypernetwork}. We also evaluate link prediction performances of approaches with respect to number of parameters. In 1vsAll or KvsAll, for each test triple $\triple{h}{r}{t}$, the score of $\triple{h}{r}{x}$ triples for all $x \in \entities$ is computed. Based on these scores, the filtered ranking $rank_t$ of the triple having $t$ is obtained. Then 
the \ac{MRR}: $\frac{1}{|\kg^{\text{test}}|} \sum_{\triple{h}{r}{t} \in \kg^{\text{test}}} \frac{1}{rank_t}$ is computed. Next, Hi@1, H@3, and H@10 are computed in literature
~\citep{dettmers2018convolutional,balavzevic2019tucker,ruffinelli2019you}. The number of parameters consists of $|\textbf{E}|, |\textbf{R}|$, and all other trainable parameters. Moreover, we illustrate our methodology with DistMult for three reasons: (1) many recent \ac{KGE} model can be seen as an effective extension of DistMult, i.e., the Hadamard product followed by an inner product of input embeddings (see Section~\ref{sec:related-work}). (2) Findings of~\citet{ruffinelli2019you} indicate that DistMult perform competitive performance provided that it is properly trained. (3) DistMult requires less floating point operations than more sophisticated recent models. Hence, it can be trained in less time~\cite{ampligraph,valeriani2020runtime}.
\subsection{Implementation Details and Reproducibility}
\label{subsec:implementation_details}
We built a hardware-agnostic knowledge graph embedding framework \footnote{\url{https://github.com/dice-group/dice-embeddings}} based on PyTorch Lightning~\cite{falcon2019pytorch} and DASK~\cite{dask}. In our experiments, the seed for the pseudo-random generator was fixed to 1. To alleviate the hardware requirements for the reproducibility of our results, we provide hyperparameter optimization, training and evaluation scripts along with pretrained models.
\section{Results}
\label{sec:results}

\subsection{Standard Link Prediction}

Table~\ref{table:umls_kinship_1vsall} reports link prediction results on benchmark datasets. 
Overall results indicate that applying \ac{KD} on embedding matrices reduces the number of required parameters, while yielding a competitive performance.
~\ac{KD}-DistMult outperforms DistMult and \ac{KD}-Rel-DistMult in 9 out of 10 metrics in Table~\ref{table:umls_kinship_1vsall} with surprisingly less parameters. 
Specifically,~\ac{KD}-DistMult outperforms DistMult and \ac{KD}-DistMult, while requiring $\mathbf{18.2 \times}$ and $\mathbf{11.4 \times}$ fewer number of parameters than DistMult and \ac{KD}-Rel-DistMult on UMLS, respectively. 
Similarly,~\ac{KD}-DistMult requires $\mathbf{14.6 \times}$ and $ \mathbf{14.2 \times}$ fewer number of parameters than DistMult and \ac{KD}-Rel-DistMult on KINSHIP, respectively. 
After observing these results, we delved into the details of training process to validate the existence of overfitting. 
To this end, we evaluated each model on the training datasets and added the MRR and Hit@N scores in Table~\ref{table:umls_kinship_1vsall}. 
\begin{table}[ht]
    \caption{Link prediction results on UMLS and KINSHIP. $|\Theta|$ denotes the number of parameters. Bold entries denote best results.}
    \small
    \scalebox{1.0}{
    \begin{tabular}{l  c c c c c c c c c c c c c c c}
      \toprule
                           &\multicolumn{5}{c}{\textbf{UMLS}}    & \multicolumn{5}{c}{\textbf{KINSHIP}}\\
                        \cmidrule(lr){2-6}  \cmidrule(lr){7-11}                       
                            &$|\Theta|$         & MRR           & @1       & @3           & @10          &$|\Theta|$   &MRR            &@1           &@3        &@10     \\
      \midrule
      DistMult               &67,915            &.517           &.441      &.536          &.659          &46,818       &.568           &.500         &.593      &.693    \\
      on training set        &                  &.995           &.992      &1.00          &1.00          &             &.919           &.876         &.952      &.991    \\
      \midrule
      \ac{KD}-Rel-DistMult   &42,619            &.531           &.432      &.584          &\textbf{.684} &45,420       &.562           &.493         &.588      &.694    \\
      on training set        &                  &.996           &.993      &.999          &1.00          &             &.914           &.867         &.953      &.992    \\
      \midrule
     \ac{KD}-DistMult        &\textbf{3,728}    &\textbf{.541}  &.447      &\textbf{.598} &\textbf{.684} &\textbf{3,200}&\textbf{.599} &\textbf{.534}&\textbf{.631}&\textbf{.709}\\  
      on training set        &                   &.814            &.704     &.904          &.989         &              &.705          &.572         &.804         &.950\\
\bottomrule
\label{table:umls_kinship_1vsall}
\end{tabular}}
\end{table}

These results suggest that (a) all models seem to suffer from overfitting and (b) increasing embedding vector size does not proportionally increase the expressiveness of the model. 
DistMult with a $\mathbf{59\%}$ more number of parameters than \ac{KD}-Rel-DistMult does not lead to a significant change in MRR and Hit@N scores on the training datasets. 
These results also highlight the importance of applying extensive extensive hyperparameter optimization and model calibration to increase generalization performances.

\subsection{Link Prediction with Model Calibration}
To observe the impact of model calibration, we retrained models with Label Smoothing and Label Relaxation (see Section~\ref{subsec:model_calibration}).
Table~\ref{table:model_calibration_umls_kinship_1vsall} shows that performances of models are improved with model calibration on UMLS, whereas performances are not increased and in some circumstances decreased. 
Importantly, model calibration seem to help more on those models that suffer greatly from the overfitting. 
For instance, models seem to perform better without model calibration on KINSHIP, where the overfitting is less severe.

\begin{table}[ht]
    \caption{Link prediction results on UMLS and KINSHIP with model calibration. 
    Rows with on training set report the performances on the training dataset. $|\Theta|$, LS, LR denote the number of parameters, Label Smoothing and Label Relaxation, respectively. Bold entries denote best results. }
    \scalebox{1}{
    \begin{tabular}{l c c c c c c c c c c c c c c c }
      \toprule
                                 &            &\multicolumn{4}{c}{\textbf{UMLS}}        &                \multicolumn{4}{c}{\textbf{KINSHIP}}\\
                                   \cmidrule(lr){2-6}  \cmidrule(lr){7-11}   
                                 &$|\Theta|$  & MRR         &@1          &@3            &@10          &$|\Theta|$   &MRR          &@1           &@3           &@10        \\
      \midrule
DistMult                         &67,915      &.517         &.441         &.536         &.658         &46,818       &.568         &.499         &.593         &.693      \\
on training set                     &            &.995         &.992         &1.00         &1.00         &             &.919         &.876         &.952         &.991          \\
with LS $\alpha=.1$     &            &.568         &.499         &.602         &.707         &             &.515         &.417         &.563         &.685      \\
on training set                     &            &.996         &.993         &1.00         &1.00         &             &.918         &.845         &.954         &.992          \\
with LS $\alpha=.2$     &            &.548         &.475         &.582         &.690         &             &.502         &.398         &.555         &.667      \\
on training set                     &            &.995         &.992         &.999         &1.00         &             &.916         &.871         &.952         &.992          \\
with LR $\alpha=.1$     &            &.552         &.436         &.630         &.723         &             &.567         &.499         &.592         &.694      \\
on training set                     &            &.995         &.992         &1.00         &1.00         &             &.919         &.875         &.952         &.992          \\
with LR $\alpha=.2$     &            &\textbf{.579}&\textbf{.487}&\textbf{.648}&.725         &             &.567         &.499         &.592         &.695      \\
on training set                     &            &.995         &.992         &1.00         &1.00         &             &.917         &.873         &.952         &.992          \\
\midrule
\ac{KD}-Rel-DistMult             &42,619      &.531         &.432         &.584         &.684         &32,946       &.556         &.487         &.580         &.689      \\
on training set                     &            &.996         &.993         &.999         &1.00         &             &.913         &.865         &.951         &.992          \\
with LS $\alpha=.1$ &    &.565         &.493         &.611         &.704         &             &.500         &.394         &.555         &.677      \\
on training set                             &    &.996         &.992         &.999         &1.00         &             &.913         &.866         &.951         &.991          \\
with LS $\alpha=.2$ &    &.592         &.531         &.617         &.704         &             &.504         &.403         &.556         &.671      \\
on training set                             &    &.994         &.990         &.999         &1.00         &             &.907         &.855         &.949         &.991          \\
with LR $\alpha=.1$&     &.543         &.444         &.602         &.692         &             &.555         &.486         &.577         &.692      \\
on training set                            &     &.996         &.993         &.999         &1.00         &             &.912         &.864         &.952         &.992          \\
with LR $\alpha=.2$&     &.562         &.476         &.600         &.718         &             &.555         &.487         &.576         &.689      \\
on training set                            &     &.993         &.999         &1.00         &1.00         &             &.912         &.863         &.952         &.992          \\
\midrule
\ac{KD}-DistMult                 &3,728       &.541         &.447         &.598         &.684         &3,200        &\textbf{.599}&\textbf{.534}&\textbf{.631}&.709      \\
on training set                     &            &.814         &.704         &.904         &.989         &             &.665         &.519         &.774         &.936      \\
with LS $\alpha=.1$ &        &.592         &.511         &.627         &.751         &             &.519         &.382         &\textbf{.621}&.705      \\
on training set                     &            &.796         &.677         &.897         &.984         &             &.640         &.491         &.745         &.928      \\
with LS $\alpha=.2$ &        &.572         &.475         &.623         &\textbf{.754}&             &.486         &.326         &.605         &.704       \\
on training set                     &            &.781         &.655         &.888         &.980         &             &.629         &.476         &.730         &.930       \\          
\bottomrule
\label{table:model_calibration_umls_kinship_1vsall}
\end{tabular}}
\end{table}

\subsection{Link Prediction under Noise}
\label{subsec:lp_under_noise}
We were interested to observe the impact of adding noisy triples into the training dataset in the link prediction task, since many real-world \acp{KG} contains noisy triples.
To this end, we add 10\% noise in the training splits and evaluate models. Table~\ref{table:noisy_umls_kinship_1vsall} suggest that 10\% noise in the input data decrease the standard DistMult model by absolute 7\% in MRR on KINSHIP, whereas performance of \ac{KD}-Rel-DistMult and \ac{KD}-DistMult are more robust against the input noise. 
Surprisingly, \ac{KD}-Rel-DistMult reaches the highest MRR, Hit@1 and Hit@3 scores throughout our experiments with 10\% noisy data. 
Bishop~\cite{bishop1995training} showed that the addition of noise to the numerical input training data lead to significant improvements in generalization performance. 
Our results signal that adding additional noise in the structured data may have the similar effect.
\begin{table}[ht]
    \caption{Link prediction results on noisy UMLS and KINSHIP. $|\Theta|$ denotes the number of parameters. Bold entries denote best results.}
    \scalebox{.85}{
    \begin{tabular}{l c c c c c c c c c c c c c c c c c}
      \toprule
                       &\multicolumn{5}{c}{\textbf{UMLS}}    & \multicolumn{5}{c}{\textbf{KINSHIP}}\\
                        \cmidrule(lr){2-6}                           \cmidrule(lr){7-11}                       
                            &$|\Theta|$   & MRR         & @1          & @3       & @10         &$|\Theta|$   &MRR       &@1        &@3        &@10       \\
      \midrule
    DistMult                &23,500       &.448         &.319         &.484      &.741         &16,200       &.523      &.452      &.538      &.665      \\
                            &28,435       &.470         &.382         &.497      &.677         &19,602       &.512      &.444      &.523      &.655      \\
                            &33,840       &.423         &.335         &.411      &.621         &23,328       &.508      &.439      &.517      &.658      \\
                            &39,715       &.518         &.418         &.556      &.741         &27,378       &.510      &.440      &.524      &.658      \\
                            &46,060       &.489         &.355         &.586      &.738         &31,752       &.512      &.438      &.529      &.668      \\
                            &52,875       &.465         &.339         &.500      &.677         &36,450       &.518      &.445      &.534      &.671      \\
                            &60,160       &.422         &.340         &.413      &.591         &41,472       &.520      &.447      &.536      &.679      \\
                            &67,915       &.485         &.396         &.497      &.699         &46,818       &.517      &.443      &.534      &.679      \\
                            &76,140       &.532         &.436         &.562      &\textbf{.746}&52,488       &.523      &.447      &.547      &.674      \\
                            &84,835       &.448         &.341         &.467      &.695         &58,482       &.529      &.453      &.551      &.684      \\
                            &94,000       &.466         &.343         &.541      &.662         &64,800       &.529      &.454      &.556      &.685      \\
      
    \midrule
    \ac{KD}-Rel-DistMult    &15,130       &.521         &.428         &.534         &.730      &11,610       &.540      &.477       &.554      &.674      \\
                            &18,205       &.550         &.467         &.578         &.690      &13,992       &.516      &.449       &.528      &.655      \\
                            &21,564       &.456         &.349         &.499         &.676      &16,596       &.507      &.442       &.514      &.644      \\
                            &25,207       &.469         &.338         &.591         &.718      &19,422       &.500      &.439       &.502      &.631      \\
                            &29,134       &.515         &.422         &.561         &.701      &22,470       &.503      &.437       &.511      &.645      \\
                            &33,345       &.580         &.500         &.617         &.701      &25,740       &.503      &.436       &.509      &.647      \\
                            &37,840       &.526         &.432         &.544         &.718      &29,232       &.506      &.438       &.515      &.644      \\
                            &42,619       &.567         &.493         &.607         &.693      &32,946       &.505      &.431       &.519      &.655      \\
                            &47,682       &.439         &.322         &.482         &.678      &36,882       &.511      &.440       &.529      &.661      \\
                            &53,029       &.554         &.490         &.574         &.671      &41,040       &.511      &.441       &.524      &.657      \\
                            &58,660       &\textbf{.583}&\textbf{.503}&\textbf{.632}&.728      &45,420       &.514      &.442       &.531      &.666      \\
    \midrule
    \ac{KD}-DistMult        &2,330        &.461        &.339          &.527         &.651      &1,600        &.562      &.487       &.592      &.699      \\
                            &2,563        &.331        &.233          &.343      &.496         &1,760        &.573      &.501       &.609      &\textbf{.703}      \\
                            &2,796        &.376        &.252          &.448      &.622         &1,920        &.567      &.497       &.599      &.696      \\
                            &3,029        &.346        &.254          &.370      &.506         &2,080        &.576      &.505       &.604      &.702      \\
                            &3,262        &.374        &.274          &.391      &.541         &2,240        &.567      &.498       &.589      &\textbf{.703}      \\
                            &3,495        &.393        &.259          &.480      &.559         &2,400        &.569      &.502       &.597      &.690      \\
                            &3,728        &.452        &.329          &.529      &.626         &2,560        &.572      &.503       &.603      &.697      \\
                            &3,961        &.374        &.276          &.385      &.655         &2,720        &.566      &.501       &.590      &.694      \\
                            &4,427        &.358        &.247          &.376      &.589         &2,880        &.578      &.509       &.607      &\textbf{.703}      \\
                            &4,194        &.545        &.438          &.596      &.744         &3,040        &.584         &.520         &.609      &.698      \\
                            &4,660        &.366        &.258          &.406      &.541         &3,200        &\textbf{.586}&\textbf{.521}&\textbf{.612}&.612      \\

    \midrule
    avg. DistMult            &55,225        &.470         &.364          &.501         &.691         &38,070      &.518      &.446       &.535      &.670      \\
    avg. \ac{KD}-Rel-DistMult&34,765        &\textbf{.524}&\textbf{.431} &\textbf{.565}&\textbf{.700}&26,850      &.510      &.443       &.521      &.653      \\
    avg. \ac{KD}-DistMult    &\textbf{3,495}&.398         &.287          &.441         &.594         &\textbf{2,400}   &\textbf{.573}&\textbf{.504}&\textbf{.601}&\textbf{.691}\\
\bottomrule
\label{table:noisy_umls_kinship_1vsall}
\end{tabular}}
\end{table}

\subsection{Parameter Analysis}
Table~\ref{table:large_umls_kinship_1vsall} and Table~\ref{table:low_umls_kinship_1vsall} report performances with a wide range of embedding vector sizes. 
Overall, our results corroborate our hypothesis, namely, as the size of embedding vectors decreases, benefits of applying \ac{KD} becomes less tangible. 
More specifically, Table~\ref{table:large_umls_kinship_1vsall} suggests that as $|\Theta|$ grows \ac{KD}-Rel-DistMult and \ac{KD}-DistMult perform quite well compared to DistMult on the both benchmark datasets. 

\begin{table}[ht]
    \caption{Link prediction results on UMLS and KINSHIP with the highest half of the the parameter sweep in the number of parameters $|\Theta|$. Bold entries denote best results.}
    \scalebox{.95}{
    \begin{tabular}{l  c c c c c c c c c c c c c c c}
      \toprule
                       &\multicolumn{5}{c}{\textbf{UMLS}}    &        \multicolumn{5}{c}{\textbf{KINSHIP}}\\
                    \cmidrule(lr){2-6}                           \cmidrule(lr){7-11} 
                       &$|\Theta|$  & MRR         &@1           &@3           &@10          &$|\Theta|$   &MRR       &@1        &@3        &@10  \\
      \midrule
      DistMult         &28,435      &.430         &.346         &.430         &.650         &19,602       &.556      &.487      &.585      &.687 \\
                       &33,840      &.439         &.357         &.455         &.545         &23,328       &.563      &.494      &.589      &.695 \\
                       &39,715      &.425         &.344         &.435         &.596         &27,378       &.562      &.493      &.587      &.695 \\
                       &46,060      &.484         &.419         &.480         &.585         &31,752       &.563      &.494      &.588      &.699 \\
                       &52,875      &.507         &.439         &.525         &.646         &36,450       &.565      &.498      &.589      &.693 \\
                       &60,160      &.459         &.363         &.487         &.646         &41,472       &.567      &.498      &.598      &.700 \\
                       &67,915      &.517         &.441         &.536         &.659         &46,818       &.568      &.500      &.593      &.693 \\
                       &76,140      &.471         &.374         &.485         &.677         &52,488       &.548      &.460      &.596      &.697 \\
                       &84,835      &.454         &.354         &.496         &.632         &58,482       &.567      &.498      &.598      &.697 \\
                       &94,000      &.436         &.353         &.439         &.587         &64,800       &.563      &.493      &.591      &.697 \\
      \midrule
 \ac{KD}-Rel-DistMult  &18,205      &.544         &\textbf{.475}&.576         &.665         &13,992       &.546      &.478      &.568      &.684  \\
                       &21,564      &.532         &.426         &.576         &\textbf{.735}&16,596       &.544      &.462      &.580      &.691  \\
                       &25,207      &.455         &.368         &.473         &.639         &19,422       &.546      &.476      &.567      &.687      \\
                       &29,134      &.477         &.373         &.508         &.661         &22,470       &.544      &.472      &.571      &.688      \\
                       &33,345      &\textbf{.525}&.452         &.548         &.642         &25,740       &.548      &.475      &.576      &.691      \\
                       &37,840      &.516         &.441         &.549         &.669         &29,232       &.553      &.484      &.578      &.689      \\
                       &42,619      &.531         &.432         &\textbf{.584}&.684         &32,946       &.556      &.487      &.580      &.689      \\
                       &47,682      &.483         &.397         &.519         &.640         &36,882       &.390      &.155      &.578      &.694      \\
                       &53,029      &\textbf{.525}&.447         &.554         &.643         &41,040       &.559      &.488      &.587      &.693      \\
                       &58,660      &.435         &.348         &.438         &.598         &45,420       &.562      &.493      &.588      &.694      \\
     \midrule
     \ac{KD}-DistMult  &2,563       &.357         &.275         &.374         &.453         &1,760        &.556      &.487      &.581      &.686      \\
                       &2,796       &.403         &.276         &.468         &.637         &1,920        &.579      &.509      &.613      &.702      \\
                       &3,029       &.354         &.271         &.375         &.492         &2,080        &.567      &.493      &.606      &.698      \\
                       &3,262       &.357         &.267         &.372         &.470         &2,240        &.577      &.511      &.605      &.701      \\
                       &3,495       &.361         &.252         &.382         &.618         &2,400        &.572      &.501      &.607      &.706      \\
                       &3,728       &.541         &.447         &.598         &.684         &2,560        &.591      &.526      &.621      &.703      \\
                       &3,961       &.407         &.282         &.474         &.645         &2,720        &.587      &.521      &.619      &.701      \\
                       &4,194       &.379         &.302         &.392         &.461         &2,880        &.591      &.526      &.620      &.708      \\
                       &4,427       &.382         &.285         &.390         &.605         &3,040        &\textbf{.599}&\textbf{.535}&.628&.707      \\
                       &4,660       &.505         &.397         &.581         &.697         &3,200        &\textbf{.599}&.534         &\textbf{.631}&\textbf{.709}\\

      \midrule
      avg. DistMult   &58,397       &.462      &.379      &.477      &.622      &40,257      &.562      &.491      &.591      &.695      \\
avg. \ac{KD}-Rel-DistMult&36,728    &\textbf{.502}&\textbf{.416}&\textbf{.532}&\textbf{.658}&28,374      &.535      &.447      &.577      &.690      \\
avg. \ac{KD}-DistMult    &\textbf{3,611}&.405      &.305      &.441      &.576      &\textbf{2,480}       &\textbf{.582}&\textbf{.514}&\textbf{.613}&\textbf{.702}      \\
\bottomrule
\label{table:large_umls_kinship_1vsall}
\end{tabular}}
\end{table}

\begin{table}[ht]
    \caption{Link prediction results on UMLS and KINSHIP with the lowest half of the the parameter sweep in the number of parameters $|\Theta|$. Bold entries denote best results.}
    \scalebox{.95}{
    \begin{tabular}{l c c c c c c c c c c c c c c c c}
      \toprule
                             &\multicolumn{5}{c}{\textbf{UMLS}}    & \multicolumn{5}{c}{\textbf{KINSHIP}}\\
\cmidrule(lr){2-6}                           \cmidrule(lr){7-11}                        
                             &$|\Theta|$  & MRR      & @1       & @3       & @10      &$|\Theta|$   &MRR       &@1        &@3        &@10    \\
      \midrule
      DistMult               &940         &.475      &.378      &.501      &.669      &648          &.450      &.291      &.559      &.686    \\
                             &2,115       &.550      &.472      &.579      &.668      &1,458        &.582      &.516      &.607      &.707    \\
                             &3,760       &.493      &.336      &.633      &.757      &2,592        &.607      &.554      &.634      &.712    \\
                             &5,875       &.584      &.498      &.626      &.749      &4,050        &.613      &.554      &.640      &.714    \\
                             &8,460       &.488      &.363      &.567      &.681      &5,832        &.612      &.555      &.631      &.709    \\
                             &11,515      &.409      &.326      &.417      &.601      &7,938        &.597      &.533      &.625      &.710    \\
                             &15,040      &.473      &.378      &.502      &.679      &10,368       &.581      &.519      &.605      &.699    \\
                             &19,035      &.548      &.476      &.590      &.709      &13,122       &.560      &.492      &.584      &.690    \\
                             &23,500      &.489      &.409      &.518      &.598      &16,200       &.549      &.480      &.570      &.683    \\
      \midrule
 \ac{KD}-Rel-DistMult        &754         &.385      &.220      &.550      &.683      &546         &.508      &.428      &.532      &.679    \\
                             &1,557       &.537      &.444      &.582      &.692      &1,152       &.382      &.137      &.581      &.694      \\
                             &2,644       &.531      &.453      &.559      &.646      &1,980       &.579      &.508      &.609      &.710      \\
                             &4,015       &.547      &.466      &.579      &.694      &3,030       &.591      &.523      &.621      &.712      \\
                             &5,670       &.457      &.367      &.487      &.635      &4,302       &.599      &.536      &.620      &.709      \\
                             &7,609       &.560      &.431      &.657      &.742      &5,796       &.592      &.529      &.615      &.709      \\
                             &9,832       &.555      &.477      &.576      &.716      &7,512       &.581      &.517      &.607      &.703      \\
                             &12,339      &.507      &.415      &.550      &.679      &9,450       &.575      &.512      &.593      &.698      \\
                             &15,130      &.565      &.503      &.586      &.687      &11,610      &.560      &.493      &.581      &.693      \\
     \midrule
     \ac{KD}-DistMult        &466         &.354      &.266      &.384      &.539      &320         &.358      &.358      &.400      &.528      \\
                             &699         &.321      &.215      &.309      &.564      &480         &.428      &.380      &.421      &.508      \\
                             &932         &.347      &.252      &.357      &.643      &640         &.492      &.422      &.505      &.643      \\
                             &1,165       &.360      &.260      &.381      &.503     &800         &.524      &.454      &.547      &.660      \\
                             &1,398       &.335      &.225      &.347      &.620     &960         &.527      &.457      &.553      &.655      \\
                             &1,631       &.375      &.225      &.493      &.651     &1,120       &.550      &.483      &.573      &.676      \\
                             &1,864       &.357      &.248      &.362      &.686     &1,280       &.575      &.503      &.607      &.708      \\
                             &2,097       &.355      &.244      &.373      &.648     &1,440       &.572      &.500      &.603      &.707      \\
                             &2,330       &.533      &.433      &.588      &.737     &1,600       &.557      &.488      &.587      &.689      \\
    \midrule
    avg. DistMult            &10,026        &.501      &.404      &.548      &.679      &6,912      &\textbf{.572}      &\textbf{.499}      &\textbf{.606}      &\textbf{.701}      \\
    avg. \ac{KD}-Rel-DistMult&6,616         &\textbf{.516}      &\textbf{.420}      &\textbf{.569}      &\textbf{.686}      &5,042      &.552      &.465      &.595      &\textbf{.701}      \\
    avg. \ac{KD}-DistMult    &\textbf{1,398}&.371      &.263      &.399      &.621      &\textbf{960}        &.509      &.449      &.534      &.642      \\
\bottomrule
\label{table:low_umls_kinship_1vsall}
\end{tabular}}
\end{table}

\section{Discussion}
\label{sec:discussion}
We conjecture that all \ac{KGE} models may benefit from an extensive hyperparameter optimization. Yet, here, we were interested in relative link prediction performances under optimizing solely the embedding vector size. 
We aimed to observe possible benefits of learning compressed embeddings in link prediction task. 
Table~\ref{table:umls_kinship_1vsall} and Table~\ref{table:low_umls_kinship_1vsall} suggest that benefits of learning compressed embeddings becomes more beneficial as the embedding vector size grows. 
Increasing the size of embedding vectors in DistMult does not decrease training loss as well as MRR and Hit@N scores on training datasets. 
This may stem from the fact that increasing the size of embeddings increases redundancy in the parameters, hence does not improve training and test performance of DistMult. 
Yet, Table \ref{table:model_calibration_umls_kinship_1vsall} shows that model calibration consistently improved generalization performances on UMLS, while performances on KINSHIP are not improved as such. 
In our experiments, performance of KD-Rel-DistMult and KD-DistMult suggest that learning compressed knowledge graph embeddings does not only lead to a competitive and sometimes superior performance but also decrease the the need of overparameterization. 
We argue that a comprehensive study including many recent state-of-the-art knowledge graph embedding models on benchmark datasets from different domains is needed to further analyse the benefits of learning compressed embeddings via KD.

\section{Conclusion}
\label{sec:conclusion}
Most \ac{KGE} models learn embeddings of entities and relations tailored towards the link prediction problem. 
Recent results signal an ever increasing predictive ability with the cost of over-parameterization and computationally complexity. 
Here, we designed a generic technique based on the Kronecker Decomposition (KD) to find a remedy for the former problem. 
Through KD, interactions within an embedding vector can be incorporated in the learning process without requiring requiring additional parameters. 
This encourages feature reuse and reduce redundancy in the embedding vectors. 
We showed that our technique can be readily applied in existing embedding models. 
Our experiments suggest that applying KD on entity and relation embeddings during training makes models more robust against overfitting and noise in input knowledge graphs. Benefits of KD becomes more tangible as the size of the input vector grows. 

\begin{acks}
This work has been supported by the German Federal Ministry of Education and Research (BMBF) within the project DAIKIRI under the grant no 01IS19085B and by the the German Federal Ministry for Economic Affairs and Energy (BMWi) within the project RAKI under the grant no 01MD19012B.
\end{acks}
\bibliographystyle{ACM-Reference-Format}
\bibliography{bibfile}
\appendix
\end{document}